\documentclass{article}
\usepackage{arxiv}

\usepackage[utf8]{inputenc} 
\usepackage[T1]{fontenc}    
\usepackage{hyperref}       
\usepackage{url}            
\usepackage{booktabs}       
\usepackage{amsfonts}       
\usepackage{nicefrac}       
\usepackage{microtype}      
\usepackage{graphicx}
\usepackage{natbib}
\usepackage{doi}
\usepackage{amsmath}

\let\ACMmaketitle=\maketitle
\renewcommand{\maketitle}{\begingroup\let\footnote=\thanks \ACMmaketitle\endgroup}

\title{Understanding Dynamic Spatio-Temporal Contexts in Long Short-Term Memory for Road Traffic Speed Prediction\footnote{This is a study we conducted while participating in the KDD Cup in 2017.}}

\date{} 					

\author{
Won Kyung Lee$^1$\thanks{equal contribution}~~~ 
Deuk Sin Kwon$^2$\footnotemark[2]~~~
So Yong Sohn$^3$\thanks{corresponding author}~~~
\smallskip 
\\
$^1$$^2$$^3$Department of Industrial Engineering, Yonsei University \\
134 Shinchon-dong, Seoul 120-749, Republic of Korea\\
$^1$wk.lee@yonsei.ac.kr, $^2$kds0281@gmail.com, $^3$sohns@yonsei.ac.kr\\
}

\renewcommand{\headeright}{}
\renewcommand{\undertitle}{}
\renewcommand{\shorttitle}{}

\hypersetup{
pdftitle={Understanding Dynamic Spatio-Temporal Contexts in Long Short-Term Memory for Road Traffic Speed Prediction},
pdfauthor={Won Kyung Lee, Deuk Sin Kwon, So Young Sohn},
pdfkeywords={Spatio-temporal dependence, Deep learning Traffic flow prediction},
}

\begin{document}
\maketitle

\begin{abstract}
	Reliable traffic flow prediction is crucial to creating intelligent transportation systems. Many big-data-based prediction approaches have been developed but they do not reflect complicated dynamic interactions between roads considering time and location. In this study, we propose a dynamically localised long short-term memory (LSTM) model that involves both spatial and temporal dependence between roads. To do so, we use a localised dynamic spatial weight matrix along with its dynamic variation. Moreover, the LSTM model can deal with sequential data with long dependency as well as complex non-linear features. Empirical results indicated superior prediction performances of the proposed model compared to two different baseline methods.
\end{abstract}

\keywords{Spatio-temporal dependence \and Deep learning \and Traffic flow prediction}

\section{Introduction}
    Reliable traffic flow prediction is an important area in the transportation field. It has great potential for various applications such as dynamic route guidance, in-vehicle information, congestion management and automatic incident-detection systems. Some recent changes in this area are the utilisation of a large quantity of real-time traffic data collected with state-of-the-art technologies.
    
    Many researchers have previously proposed parametric time-series models to predict traffic flow. Typical traffic prediction models were based on an autoregressive integrated moving average (ARIMA) model. For instance, a seasonal ARIMA (SARIMA) model or an ARIMA with exogenous variables (ARIMAX) model were proposed. Such models usually assume a simple linear structure. However, traffic flow changes because of complicated interactions among roads and that aspect cannot be fully considered via such a simple linear structure in the model \cite{vlahogianni2007statistical}. For this reason, many non-parametric methods have also been proposed. 
    
    In particular, neural network models show superior predictability among various non-parametric models. Furthermore, deep neural network models have been actively investigated as big traffic data became available. Many extended models have been proposed, from typical fully connected feed-forward neural network models with random weight initialisation \cite{huang2014deep} to numerous pre-training methods such as the restricted Boltzmann machine \cite{siripanpornchana2016travel} or stacked autoencoder \cite{lv2014traffic}. However, even though those models were constructed with deep architecture, the fully connected feed-forward structure is not enough to consider sequential real-time traffic data for traffic flow prediction. Additionally, the proposed models did not consider potential spatial dependence between roads.
    
    On the other hand, some models that deal with spatio-temporal dependence have been suggested to predict traffic flow \cite{min2011real, wang2016traffic,zhao2017lstm}. The common feature of these models when we predict traffic flow in a specific road are its previous traffic records and traffic flow of neighbouring roads. However, such dependence structure also changes with time and place \cite{cheng2012spatio}. While this localised dynamic spatio-temporal dependence in traffic flow has been considered in a simple linear model \cite{cheng2014dynamic}, no effort has been made to include this dependence in more complex models.
    
    Therefore, in this study, we propose a deep learning model, a dynamically localised long short-term memory (LSTM) network, which considers dynamic spatial dependence for traffic flow prediction and its local variation.

\section{Literature Review}
    Existing methods for traffic flow prediction can be categorised into parametric and non-parametric methods. In this section, we review the literature related to both categories and also discuss the research gaps in the models for spatio-temporal dependence. 
    
    \subsection{Parametric Models}
        Many studies used the ARIMA model, a representative time-series model for travel time prediction to reflect highly dependent traffic flows over time. Some researchers \cite{ahmed1979analysis,levin1980forecasting} introduced the ARIMA model as an effort to integrate probabilistic traffic mechanisms for prediction. They all used the Box–Jenkins time series ARIMA model. However, later on, modified ARIMA models were proposed for enhanced prediction performance. For example, \citet{williams2001multivariate} improved the predictive performance of the ARIMA model using exogenous inputs in the model (ARIMAX). \citet{williams2003modeling} used the seasonal ARIMA (SARIMA) for travel time prediction for periodic traffic flow. In addition to the ARIMA model, several other parametric methods, such as the Markov chain \cite{qi2014hidden} and the Bayesian network \cite{wang2014new}, have been suggested. They commonly use a stochastic approach. However, most of the proposed parametric methods are limited since they only deal with a simple linear structure, not displaying a highly accurate prediction performance.

    \subsection{Non-parametric Models}
    Non-parametric models have been actively investigated to consider complex relationships, including non-linear ones, between the variables. For example, non-parametric regression analysis based on K-nearest neighbour \cite{chang2012dynamic} and regression analysis based on kernel trick \cite{haworth2012non} were proposed to predict traffic flow. Moreover, several traffic prediction models that have been suggested are based on artificial neural networks and have involved multi-layer perceptrons \cite{smith1994short}, time-delayed neural networks \cite{lingras2000traffic} and recurrent neural networks (RNN) \cite{dia2001object, ma2015large}.
    
    Overall, neural-network-based models have shown better performance than parametric models in recent literature \cite{cai2016spatiotemporal}. In fact, many researches have applied deep neural network models to predict traffic flows, owing to recent developments in training of deep neural networks \cite{ma2015long}. As we utilise deep neural networks in this study, important features that have neither been explained by nor found in transportation-based theory can be found and considered in the models reported by \citet{lv2014traffic}. Such aspects led to rapid development in terms of prediction performance. However, even though the neural network models have deep structure, they still need to consider spatio-temporal dependence between roads. 

    \subsection{Spatio-temporal Models}
    When observations consider spatio-temporal dependence, reflecting such dependence with appropriate quantitative approaches is necessary \cite{hackney2007predicting}. Specifically, travel time, speed and traffic volume have a high spatio-temporal autocorrelation. Some studies therefore investigated spatio-temporal dependence and reflected it in traffic flow prediction models \cite{liu2011discovering}. 
    
    First, a space-time ARIMA (STARIMA) model was proposed to consider spatio-temporal dependence \cite{min2011real}. Later, \citet{cheng2014dynamic} suggested a localised STARIMA that applies a dynamic spatial weight matrix because the level of spatio-temporal dependence can vary over time and region. In their research, authors defined spatial neighbours of a road as the roads that can be reached based on road velocity. Moreover, the spatial weights were determined based on the relative velocity. However, in network analysis, second or higher-order neighbours would have indirect effects while first-order neighbours have a direct impact. Therefore, the procedure to determine dynamic spatial weights can be modified by employing distance-decaying weight. Algorithmic improvement is also necessary to consider a non-linear structure. 
    
    Some neural-network-based models were established considering the spatio-temporal dependence. Two types of neural network models, RNN and convolutional neural network (CNN), are successful in various research fields and have become popular when considering spatial or temporal dependence in a prediction task. Transportation researchers also began to employ different network structures for traffic flow prediction. For instance, \citet{van2005accurate} exploited a state-space neural network that utilises spatio-temporal input–output mapping and has an RNN structure. Some authors integrated the RNN and CNN structures and developed an error-feedback RCNN (eRCNN) that reflects both spatial and temporal dependence among roads for traffic speed prediction \cite{wang2016traffic}. However, these works did not assume that the degree of auto-correlation could change depending on time or location.
    
    Such limitation was overcome in other studies. \citet{wang2016space} proposed a space-time delay neural network (STDNN) to consider spatio-temporal autocorrelation as well as its local variation in traffic flow prediction. However, the proposed STDNN has only a single hidden layer so that it is not adequate for scaling into deep architecture due to an over-fitting problem. On the other hand, \citet{zhao2017lstm} constructed an origin destination correlation (ODC) matrix, which embodies spatio-temporal dependence among roads, and used the matrix as an input for LSTM networks. Still, this model needs to calculate the correlation for every pair of roads to build the ODC matrix because the authors assume all of the roads directly affect a specific road based only on the correlation. Thus, when many roads are involved, the proposed method can be inefficient and inappropriate as it does not consider the distance between roads or the road structure.

\section{Methodology}   
    Traffic flow prediction is not easy because there are complex interactions between roads and those interactions change over time and location. Therefore, not only a history of traffic flow in the single road but also the data recorded nearby should be utilised. In this context, we developed a traffic flow prediction model using an LSTM network to consider the complex interaction in terms of temporal dependence. This LSTM network for traffic prediction covers both long-term and short-term changes caused by sudden traffic accidents or other events. At the same time, our model utilises a localised dynamic spatial weight matrix to reflect spatial dependence.    
    
    \begin{figure}
    	\centering
    	\includegraphics[width=10cm, height=7cm]{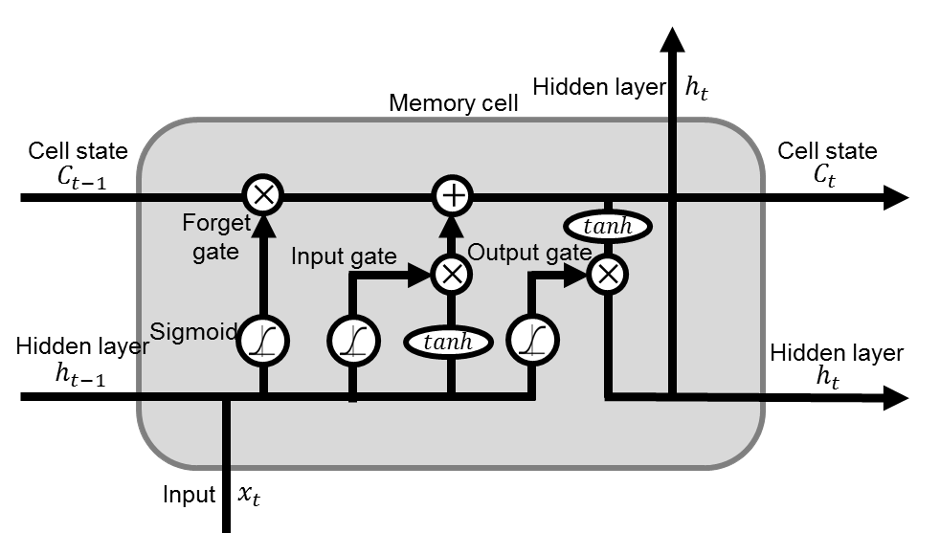}
    	\caption{Structure of the LSTM network}
    	\label{fig:fig1}
    \end{figure}
    
    \subsection{Localised Dynamic Spatial Weight Matrix}
    The travel time of a particular road link can be related to its travel time in the past, but it is also closely related to traffic situations of the neighbouring roads. In terms of the degree of relationship, the nearest neighbours can be intuitively thought of as having the closest relationship. The spatial weight matrix is often used as a way to reflect space dependence in spatial research. The neighbours geographically adjacent to the area are called first-order neighbours and the neighbours of neighbours are called second-order neighbours, depicting these relationships as matrices that represent spatial dependence in the analytical model. However, in traffic flow prediction, the spatial dependence relationship may be different according to the road link and may also change with the time zone. For example, if the average speed on a road is generally slow, the range of influence of traffic flow on other roads, e.g. a neighbouring road, will be narrow. On the other hand, when the average speed of some roads is high, the range of reachable neighbouring roads will be wide. Therefore, it is not appropriate to apply the spatial dependency relationship equally across the globe; but it should be able to reflect the spatial dependence varying over time.
    
    \begin{figure}
    	\centering
    	\includegraphics[width=15cm, height=13cm]{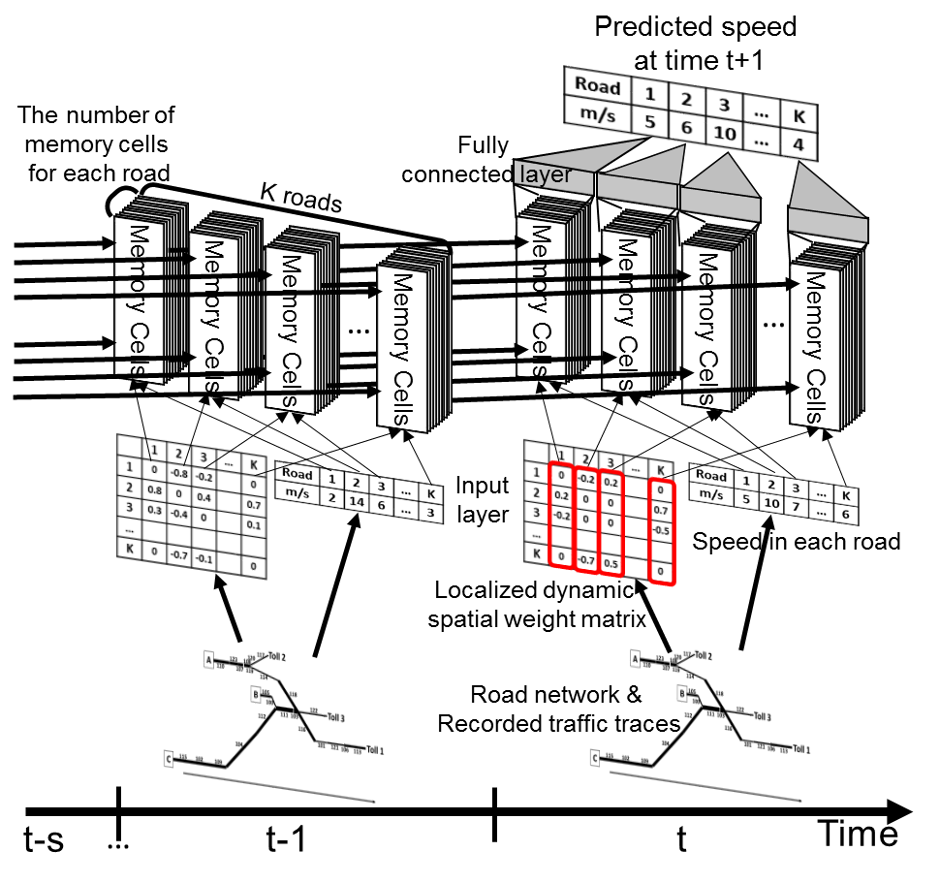}
    	\caption{Proposed framework based on the dynamically localised LSTM model to predict the speed at time $t$+1}
    	\label{fig:fig2}
    \end{figure}
    
    To reflect changes in the spatial dependence relationship over time, we refer to the dynamic spatial weight matrix proposed by \citet{cheng2014dynamic}. If we are to express the spatial dependence relation of the roads $k$ and $l$ of an area with $n$ road links in total, as a dynamic spatial weight matrix $M$, the matrix at time t $M_{t}$ with elements ($k$,$l$) is defined as follows:
    
    \begin{center} $M_{t}$ ($k$,$l$) = \textrm{rate of change at road}\ $l$\ \textrm{caused by road}\ $k$ 
    \end{center}
    
    \begin{equation}
        \textrm{, where k = }
        \\ \left \{ \begin{array}{cc} 
        (\cfrac{v_l - v_k}{v_l}) \cdot (\cfrac{1}{d_{lk}}) \  \textrm{,when road $k$ is behind load $l$} \\
        (\cfrac{v_k - v_l}{v_l}) \cdot (\cfrac{1}{d_{lk}}) \  \textrm{,when road $k$ is ahead of load $l$} \\
         0 \textrm{,when road $k$ is out of the coverage of load $l$} \end{array} \right \}
    \end{equation}
    Let $v_l$ and  $v_k$ be respectively the speeds at roads $l$ and $k$, and $d_{lk}$ be the distance between the two roads. The coverage of road l means the area within the distance that can be reached within the unit time with speed $v_l$ at the road. The dynamic spatial weight matrix, as described above, reflects the structure in which the coverage range is determined according to the speed of the road and time; the rate of change in traffic flow dynamically changes as time progresses. Based on the matrix (1), we can use the information in the $l$-th column of $M_{t}$ to predict the traffic flow of road l after $\Delta{t}$ of time change at time $t$

    \subsection{LSTM Network-Based Analytic Framework for Traffic Speed Prediction}
    An LSTM network is a special type of RNN as it deals with time series input and has a recurrent structure. The main characteristic of the LSTM network, compared to other types of RNN, is the capability of both short and long dependency in input data. Owing to this feature \cite{pascanu2013difficulty}, the LSTM network has been applied in various fields, such as natural language processing \cite{wang2015learning} and speech recognition \cite{graves2013speech}. 
        
    In the LSTM network, there is a recurrent memory module, which comprises cell states and input-, output- and forgetting-gates, as shown in Figure  \ref{fig:fig1}. According to the figure, each memory cell receives the cell state, the hidden layer from the previous time step, as well as the current input data. Then, the memory cell output is a cell state and a hidden layer for the next time step or a higher level of layer. A detailed explanation can be found in \citet{hochreiter1997long}.

    In addition, the input and output layer may be explained as follows. Let us assume that the task is to predict an average travel speed at road $l$ after time $t$. Then, we feed the history of travel speed at road l as well as the column total of $l$-th column of the localised dynamic spatial weight matrix into the input layer and insert the travel speed after time t at that road into the output layer. In particular, the column total of $l$-th column of the matrix defined in section 3.1 implies the rate of change of travel speed in the near future at road l caused by the other roads. These procedures are summarised in Figure \ref{fig:fig2}. As time progresses, newly observed traffic conditions, the spatial weight matrix based on the observation and the features that were learned in the previous time steps are utilised to update the weights of the LSTM network. In our framework, we assume there are $k$ roads and $s$ is the time window to look back.

    \begin{figure}[b]
    	\centering
    	\includegraphics[width=11cm, height=5cm]{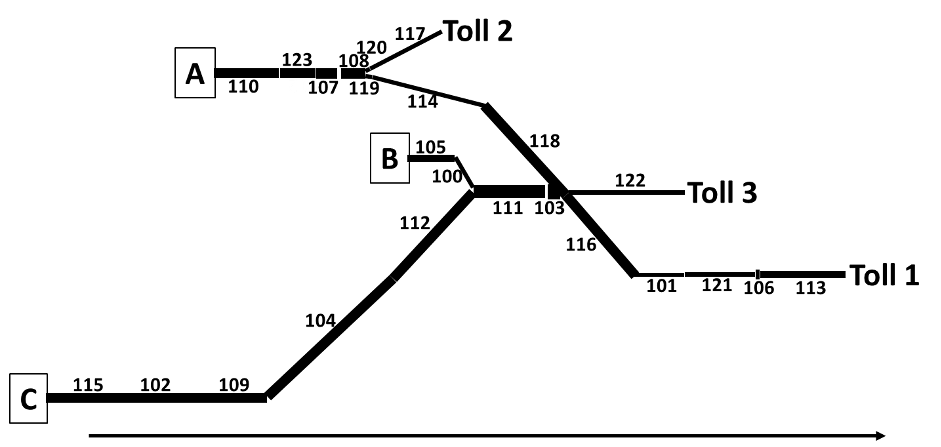}
    	\caption{Diagram of the highway investigated. Each line segment describes the corresponding road segment; the length and thickness of the line are proportional to the length of the road and the number of lanes in the road. There are six travel routes that consist of one-way roads (A $\xrightarrow[]{}$ Toll2, A $\xrightarrow[]{}$ Toll3, B $\xrightarrow[]{}$ Toll1, B $\xrightarrow[]{}$ Toll3, C $\xrightarrow[]{}$ Toll1, C $\xrightarrow[]{}$ Toll3)}
    	\label{fig:fig3}
    \end{figure}
\section{Experimental} 
    \subsection{Data Description}
    The dataset used in this study is obtained from the Knowledge Discovery and Data Mining (KDD) Cup (2017) \cite{KDD}. It comprises the travel histories of some vehicles at a highway in China. As shown in Figure \ref{fig:fig3}, there are 24 road segments, which consist of six types of travel routes from the starting point to the toll destination. The time at the entrance and exit were recorded for each road segment. Moreover, the dataset covers the traffic histories for 97 days, from 19 July to 24 October, 2016. About 80\% of the dataset, which covers data from 19 July to 3 October, 2016, is considered as the train dataset; the other is used as the test set. As there were two holidays, Moon festival (15 Sept.–17 Sept.) and Chinese national holidays (1 Oct.–7 Oct.), the train or test set include traffic data in holidays.

    In this study, the average travel speed at each road after time unit t is predicted by our proposed method. As we know both the time that vehicles enter to and exit from each road, the average travel speed can be obtained for every time period. The obtained average travel speed at each road is utilised as an input or output for our traffic prediction model.
    
    \subsection{Model Description}
    In this study, we compare our proposed LSTM model, which utilises localised dynamic spatial weight matrix, with two different baseline methods. To predict the average travel speed at a specific road after time t, the first baseline method, investigated by \citet{ma2015long}, uses only the history of the average travel speed at the specific road. Such an approach considers only temporal dependence on the travel speed to predict the future travel speed and has been widely used in previous research on time-series analysis. 
    
    The second baseline method utilises the traffic histories in all of the roads instead of the history in only a specific road. This method considers both temporal and spatial dependence. For spatial dependence, some researchers might determine the coverage of neighbours on their own, but the coverage in this baseline method can be determined via data-driven methodology, i.e. inferring weight parameters of a neural network. Through this method, the determined neighbour’s structure is used to predict the average travel speed of a specific road after time unit $t$.
    
    Moreover, the travel speed after 1 min for each road was experimentally predicted using both baseline models and our proposed model. In fact, different time units have been selected for the prediction horizon, e.g. 1, 2, 5 or 15 min \cite{vlahogianni2014short}. Since we utilise a deep-learning-based method, we set a rather short time unit to make a large scale of training dataset in order to avoid the over-fitting problem. Furthermore, short-term unit prediction is appropriate when considering a dynamic locality of spatial neighbours of a road that would be determined based on the road velocity. The models in this experiment predict the velocity 1 min ahead; they can, however, be easily extended to predict traffic conditions for longer times by feeding the predicted values into the models. 
    
    In addition, the same setting of hyper-parameters was used in the experiment for all the models. We use 10 LSTM cells for each method; every input data are processed via the min-max normalisation method. Moreover, the size of the time window, $s$, is set to 10 min. Besides, the loss function is defined as the Mean Absolute Percentage Error (MAPE), which is solved by the AdaDelta optimiser \cite{zeiler2012adadelta}, where the batch size is 32. In our experiment, we stopped the iteration when the training loss decreases but the validation loss increases. 

\section{Result}
\begin{figure}
    	\centering
    	\includegraphics[width=15cm, height=11cm]{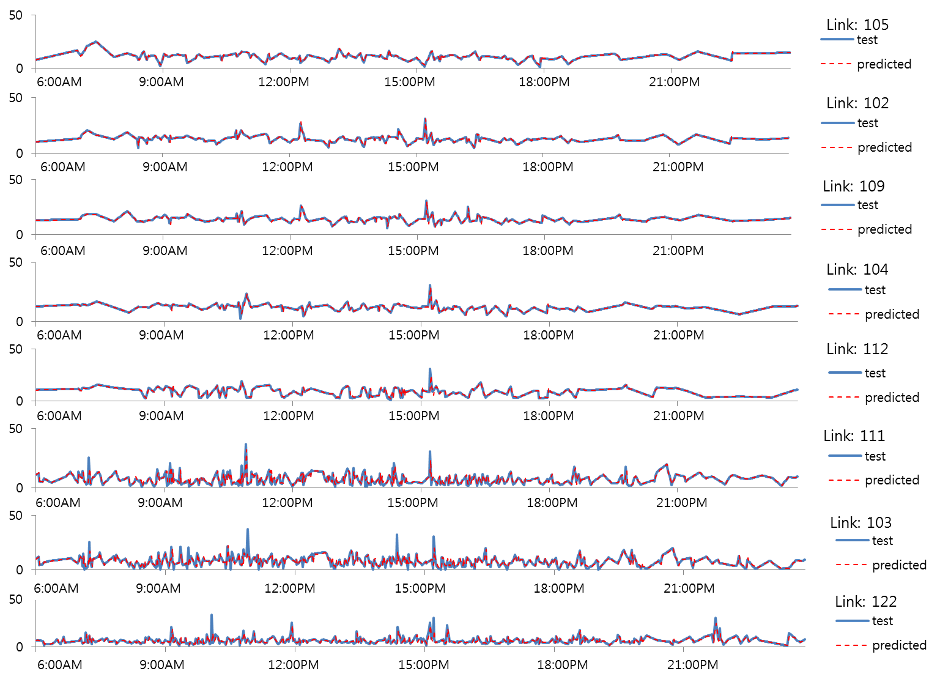}
    	\caption{Predicted travel speeds using the proposed LSTM model along with the real value for the eight roads that compose the course ‘C → Toll 3’}
    	\label{fig:fig4}
    \end{figure}

    In this section, we compare the results of the three different LSTM models. Table 1 shows the MAPE in each course that consists of several roads. As seen in Table \ref{Tab:table1}, our proposed model, which utilises localised dynamic spatial weight matrix, provides the lower value of MAPE compared to the other baseline models. Additionally, the values predicted with our model are shown in Figure \ref{fig:fig4} for a period from 6 AM to 12 AM (midnight) on 18 October, 2016. In Figure 4, the predicted travel speed and the real value for each road that composes the course ‘C → Toll 3’ are very similar to each other.

    \begin{table}[b]
        \begin{center}
		\renewcommand{\arraystretch}{1.2}
        \caption{Prediction results of the average travel speed in each course: the MAPE in testing data for the period from 3 to 23 October, 2016.}
            \begin{tabular}{c|c|c|c|c|c|c}
            \toprule	           
            Method          & C → Toll1 & C → Toll3 & C → Toll1 & B → Toll3 & A → Toll2 & A → Toll3 \\
            \hline
            Baseline Model1 & 10.141 & 12.1205 & 13.421  & 19.211  & 19.418  & 15.698\\
            \hline
            Baseline Model2 & 10.342 & 12.144 & 13.6542  & 19.186  & 18.820  & 15.578\\
            \hline
            Proposed Model  & 10.008 & 11.857  & 13.219  & 18.747  & 18.845  & 15.264\\
            \hline
            \begin{tabular}[c]{@{}l@{}}The difference in the MAPE \\ (the proposed model \textit{vs} \\ the based baseline model)\end{tabular} &
              0.133 &
              0.262 &
              0.202 &
              0.438 &
              -0.025 &
              0.313 \\
              \bottomrule	
            \end{tabular}
            \label{Tab:table1}
        \end{center}
    \end{table}
    
    Furthermore, we tested traffic prediction for unusual traffic patterns. As mentioned in section 4.1, our dataset also includes traffic during holidays. Particularly, our test data set covers some of the Chinese National holidays from 3 to 7 October. Traffic during these holidays is deviated from usual traffic patterns. Applying the trained models to the festival period data, test results are obtained as MAPE (Table \ref{Tab:table2}). The overall MAPE increased for these periods. However, the difference in MAPE between our localised LSTM model and the best-performance baseline method increased in five of the six courses. This implies that our model is more robust than the baseline methods, reflecting the advantage of the consideration of the dynamic spatial weight matrix. 
    
    \begin{table}[t]
        \begin{center}
		\renewcommand{\arraystretch}{1.2}
        \caption{Prediction results of the average travel speed for each course: MAPE in test data, which covers parts of the Chinese National holidays (3 to 7 October).}
            \begin{tabular}{c|c|c|c|c|c|c}
            \toprule
            Method          & C → Toll1 & C → Toll3 & C → Toll1 & B → Toll3 & A → Toll2 & A → Toll3 \\
            \hline
            Baseline Model1 & 13.267 & 15.456 & 17.665  & 24.687  & 17.884  & 15.698\\
            \hline
            Baseline Model2 & 13.457 & 15.406 & 17.881  & 23.539  & 22.029  & 17.633\\
            \hline
            Proposed Model  & 13.033 & 15.081  & 24.048  & 22.318  & 21.318  & 17.633\\
            \hline
            \begin{tabular}[c]{@{}l@{}}The difference in the MAPE \\ (the proposed model \textit{vs} \\ the based baseline model)\end{tabular} &
              0.234 &
              0.324 &
              0.333 &
              0.490 &
              -0.289 &
              0.005 \\
            \bottomrule	
            \end{tabular}
            \label{Tab:table2}
        \end{center}
    \end{table}

\section{Conclusions}
    In this study, we proposed a deep-learning-based travel speed prediction model: a dynamically localised LSTM model, which utilises a localised dynamic spatial weight matrix that captures the spatio-temporal dependence between roads and both its local and dynamic variation. This matrix considers the principles of dynamics in road traffic based on relative velocity. We used this matrix in an LSTM model that deals with long sequence input data to predict the travel speed in the near future. When comparing this model to other baseline models for a travel speed prediction task at a highway, the MAPE results imply that our proposed model is superior to the others in terms of accuracy and robustness under unusual traffic patterns caused by different events.
    
    Our localised LSTM model could contribute to several applications. For example, as it improves accuracy as well as robustness when predicting travel speed for each road, this may be used to better infer travel time. Moreover, such accurate and robust prediction can contribute to making decisions such as route choices for trip planning or en-route navigation \cite{fei2011bayesian}.
    
    The model proposed in this research has however some limitations and needs to consider additional aspects in further studies. First, the experiments were conducted using the traffic records of highways that constitute a simple road structure. Future research could investigate urban roads where many intersections exist. In that case, it would be necessary to consider the effect of traffic signals on forecasting. Another aspect to be improved is generalisation. When applying this method to other regions, we need a large scale of train data, which would require a lot of time; our proposed model learned several complex features via a deep-learning-based approach. Thus, we still need methods to extract the general features that could be applied globally; for example. transfer learning, which transfers and utilises knowledge from source domains to target domains.

\section*{Acknowledgements}
    This research, 'Geospatial Big Data Management, Analysis and Service Platform Technology Development', was supported by the MOLIT (The Ministry of Land, Infrastructure and Transport), Korea, under the national spatial information research program supervised by the KAIA (Korea Agency for Infrastructure Technology Advancement)"(17NSIP-B081011-04)

\clearpage
\bibliographystyle{unsrtnat}
\bibliography{references}  






\end{document}